\documentclass[sigconf]{aamas}

\usepackage{balance} 

\setcopyright{ifaamas}
\acmConference[AAMAS '23]{Proc.\@ of the 22nd International Conference
on Autonomous Agents and Multiagent Systems (AAMAS 2023)}{May 29 -- June 2, 2023}
{London, United Kingdom}{A.~Ricci, W.~Yeoh, N.~Agmon, B.~An (eds.)}
\copyrightyear{2023}
\acmYear{2023}
\acmDOI{}
\acmPrice{}
\acmISBN{}

\acmSubmissionID{803}

\title{Adaptive Value Decomposition with Greedy Marginal Contribution Computation for Cooperative Multi-Agent Reinforcement Learning}

\author{Shanqi Liu}
\affiliation{
  \institution{Zhejiang University}
  \city{Hangzhou}
  \country{China}}
\email{shanqiliu@zju.edu.cn}

\author{Yujing Hu*}
\affiliation{
  \institution{NetEase Fuxi AI Lab}
  \city{Hangzhou}
  \country{China}}
\email{huyujing@corp.netease.com}

\author{Runze Wu}
\affiliation{
    \institution{NetEase Fuxi AI Lab}
    \city{Hangzhou}
    \country{China}}
\email{wurunze1@corp.netease.com}

\author{Dong Xing}
\affiliation{
    \institution{Zhejiang University}
    \city{Hangzhou}
    \country{China}}
\email{dongxing@zju.edu.cn}

\author{Yu Xiong}
\affiliation{
    \institution{NetEase Fuxi AI Lab}
    \city{Hangzhou}
    \country{China}}
\email{xiongyu1@corp.netease.com}

\author{Changjie Fan}
\affiliation{
    \institution{NetEase Fuxi AI Lab}
    \city{Hangzhou}
    \country{China}}
\email{fanchangjie@corp.netease.com}

\author{Kun	Kuang}
\affiliation{
    \institution{Zhejiang University}
    \city{Hangzhou}
    \country{China}}
\email{kunkuang@zju.edu.cn}

\author{Yong Liu*}
\affiliation{
    \institution{Zhejiang University}
    \city{Hangzhou}
    \country{China}}
\email{yongliu@iipc.zju.edu.cn}

\begin{abstract}
    Real-world cooperation often requires intensive coordination among agents simultaneously. This task has been extensively studied within the framework of cooperative multi-agent reinforcement learning (MARL), and value decomposition methods are among those cutting-edge solutions. However, traditional methods that learn the value function as a monotonic mixing of per-agent utilities cannot solve the tasks with non-monotonic returns. This hinders their application in generic scenarios. Recent methods tackle this problem from the perspective of implicit credit assignment by learning value functions with complete expressiveness or using additional structures to improve cooperation. However, they are either difficult to learn due to large joint action spaces or insufficient to capture the complicated interactions among agents which are essential to solving tasks with non-monotonic returns. Moreover, applications in real-world scenarios usually require policies to be interpretable, but interpretability is limited in the implicit credit assignment methods. To address these problems, we propose a novel explicit credit assignment method to address the non-monotonic problem. Our method, Adaptive Value decomposition with Greedy Marginal contribution (AVGM), is based on an adaptive value decomposition that learns the cooperative value of a group of dynamically changing agents. We first illustrate that the proposed value decomposition can consider the complicated interactions among agents and is feasible to learn in large-scale scenarios. Then, our method uses a greedy marginal contribution computed from the value decomposition as an individual credit to incentivize agents to learn the optimal cooperative policy. We further extend the module with an action encoder to guarantee the linear time complexity for computing the greedy marginal contribution. Experimental results demonstrate that our method achieves significant performance improvements in several non-monotonic domains. Besides, we showcase that our model maintains a good sense of interpretability and rationality. This suggests our model can be applied to scenarios with more realistic demands. 
\end{abstract}


\newcommand{\BibTeX}{\rm B\kern-.05em{\sc i\kern-.025em b}\kern-.08em\TeX}

\begin{document}

\pagestyle{fancy}
\fancyhead{}


\maketitle
\renewcommand{\thefootnote}{\fnsymbol{footnote}}
\footnotetext[1]{The corresponding authors.}
\section{Introduction}
Many real-world tasks like bimanual manipulation \cite{lee2013relative}, autonomous driving \cite{liu2020moving} and swarms \cite{huttenrauch2017guided} require the cooperation of multiple agents. 
Especially, in many scenarios agents are expected to choose the optimal actions simultaneously to complete the common goal, such as the bimanual lifting task requiring dual arms to lift the object simultaneously \cite{bonitz1996internal, caccavale2008six}. 
Learning cooperative policies in these tasks remains challenging as agents' joint action spaces grow exponentially with the number of agents and performing reward decomposition, known as credit assignment, is challenging due to complex interactions among agents. 
Cooperative multi-agent reinforcement learning (MARL) has been broadly used to learn effective behaviors in such tasks from agents' experiences. 
Currently, a popular paradigm for value-based cooperative MARL is centralized training with decentralized execution (CTDE), and representative methods include VDN and QMIX \cite{sunehag2017value,rashid2018qmix}. These methods learn the centralized value functions as monotonic factorizations of each agent's utility function and enable decentralized execution by maximizing each agent's corresponding utility function, known as the Individual Global Maximum (IGM) principle \cite{hostallero2019learning}. 
Despite the fact that these methods have successfully solved tasks with monotonic payoff matrices optimally \cite{hu2021rethinking}, they fail in tasks with non-monotonic payoff matrices. As the monotonic value function restricts the value function to sub-optimal value approximations in environments with non-monotonic payoffs \cite{wang2020qplex,son2019qtran}, they cannot represent the policy that an agent's optimal action depends on actions from other agents. 
This problem, known as the relative overgeneralization \cite{panait2006biasing}, prevents the agents from solving tasks such as bimanual lifting. 

To address this problem, recent works either aim to learn value functions with complete expressiveness capacity \cite{wang2020qplex} or use auxiliary approaches such as placing more importance on better joint actions to find the optimal cooperative policy \cite{rashid2020weighted}. However, they still face other problems. On the one hand, learning complete expressiveness is difficult since the joint action spaces are exponentially related to the number of agents. On the other hand, using auxiliary approaches to improve cooperation cannot capture the complicated interactions among agents and lacks theoretical optimal guarantees \cite{gupta2021uneven} (Section \ref{motivation} details the reasons for the problems). 
Furthermore, current methods are usually implicit credit assignment methods, which learn the credit assignment by using a mixing network. Compared to the explicit credit assignment methods based on Shapley Value \cite{shapley201617} or marginal contribution, implicit methods lack interpretability for the distributed credit, which hinders their application in scenarios where security is an essential property. 

In this work, we propose a novel explicit credit assignment method that solves the non-monotonic problem by using the greedy marginal contribution of each agent. 
First, we propose an adaptive value decomposition method that models individual utility as a distribution over actions of other observable agents. The value decomposition enables agents to learn the individual utility that can take into account the interactions with others. We illustrate that such an individual utility can represent non-monotonic payoff matrices correctly and we further investigate why previous methods that model the utility as a specific value fail in the non-monotonic settings in Section \ref{motivation}.
Furthermore, since the utility merely models interactions among observable agents, the learning complexity does not increase exponentially with the number of agents as in methods that learn complete expressiveness. 
However, although such individual utilities can describe the payoff matrices correctly, we cannot directly use it as policy's value function as it requires other agents' actions. 
To address this problem, our method optimizes each agent's decentralized policy by learning a greedy marginal contribution from the adaptive value decomposition, which calculates the optimal marginal contribution to avoid the requirement to know the other agents' actions for policy computation. 
We illustrate that optimizing the greedy marginal contribution can achieve the cooperative policy by avoiding taking actions that lead to mis-coordination. 
This also enhances exploration as the number of samples taking cooperative actions increases. 
Furthermore, since the computation time complexity of maximizing the marginal contribution is exponentially related to the number of observable agents, we propose an action encoder that maps actions into a condensed latent space to guarantee a linear searching time for the maximization. This reduces computational burden and makes the learning process practical. 
Finally, our method also increases the adaptability of learned policy to cooperate with scalable agents as our method learns the values of actions cooperating with dynamically changing observable agents. 
We evaluate our method against several state-of-the-art baselines in non-monotonic robotics tasks that require simultaneous coordination amongst agents. The result shows that our method achieves significant improvements in asymptotic performance and efficiency. 
Moreover, we showcase the interpretability of our method that current methods lack.

\section{Related Work}

Many general deep MARL methods have been used in complex multi-agent environments, including COMA \cite{foerster2018counterfactual}, MADDPG \cite{lowe2017multi}, etc.
Currently, CTDE forms the de facto mainstream paradigm in cooperative MARL \cite{lowe2017multi,iqbal2019actor}. 
Regarding the details, VDN \cite{sunehag2017value} learns the joint-action Q-values by factoring them as the sum of each agent’s utility. QMIX \cite{rashid2018qmix} extends VDN to allow the joint action Q-value to be a monotonic combination of each agent’s utility that can vary depending on the global state.
However, the monotonic constraints on the joint action-values introduced by QMIX and related QMIX-based variant methods result in provably poor exploration and relative overgeneralization \cite{panait2006biasing}. 
To address this problem, QPLEX \cite{wang2020qplex} and QTRAN \cite{son2019qtran} aim to learn value functions with complete expressiveness capacity. However, reports are that they perform poorly when being used in practice \cite{gupta2021uneven,wan2021greedy}. This is because learning the complete expressiveness is impractical in complicated MARL tasks due to the challenging exploration in large joint action spaces. 
In contrast, we restrict the modeling of agent interaction value to be within pairs of potentially cooperating agents. This prevents the complexity of learning from growing exponentially along with the size of joint action spaces. Moreover, our method also enhances exploration by increasing the probability of sampling cooperative actions. 
There also exist methods using auxiliary structures to solve the problem. For instance, MAVEN \cite{maven} hybridises value and policy-based methods by introducing a latent space for hierarchical control. This allows MAVEN to achieve committed, temporally extended exploration. 
Weighted QMIX \cite{rashid2020weighted} is based on QMIX and rectifies the suboptimality by introducing weights to place more importance on the better joint actions. 
UneVEn \cite{gupta2021uneven} learns a set of related tasks simultaneously with a linear decomposition of universal successor features. 
However, these methods cannot estimate the value of actions considering the changes in other agents' actions and usually fail in environments with extreme rewards. On the contrary, our method can estimate the contribution of each agent according to the utility considering the interactions of all cooperative agents which is shown to work in all non-monotonic environments.
Moreover, all the methods mentioned above belong to implicit credit assignment methods, which lack interpretability for the distributed credits. 

Our work is also related to credit assignment. Compared to implicit credit assignment methods, explicit methods attribute each agent's contribution to being at least provably locally optimal \cite{kinnear1994advances}. COMA \cite{foerster2018counterfactual} utilizes a counterfactual advantage to learn the value function. QPD \cite{yang2020q} designs a multi-channel mixer critic and leverages integrated gradients to distribute credits along paths. However, they use monotonic value functions and ignore the complex interactions between agents, which are critical for cooperation in non-monotonic tasks. To address this, SQDDPG \cite{wang2020shapley} and Shapley \cite{li2021shapley} use Shapley Value \cite{shapley201617} to estimate the complex interactions between agents. Shapley Value originates from cooperative game theory and is able to distribute benefits reasonably by estimating the contribution of participating agents.
However, SQDDPG \cite{wang2020shapley} assumes that the agents take actions sequentially, which does not hold in many scenarios. Shapley \cite{li2021shapley} uses a single centralized critic network to estimate Shapley Value, which is infeasible when the number of agents is large. Moreover, these methods can only get approximated Shapley Value as calculating the Shapley Value involves exponential time complexity \cite{wang2020shapley}.

\section{Motivation}
\label{motivation}
\begin{figure}{t}
    \centering
    \includegraphics[width=0.9\linewidth]{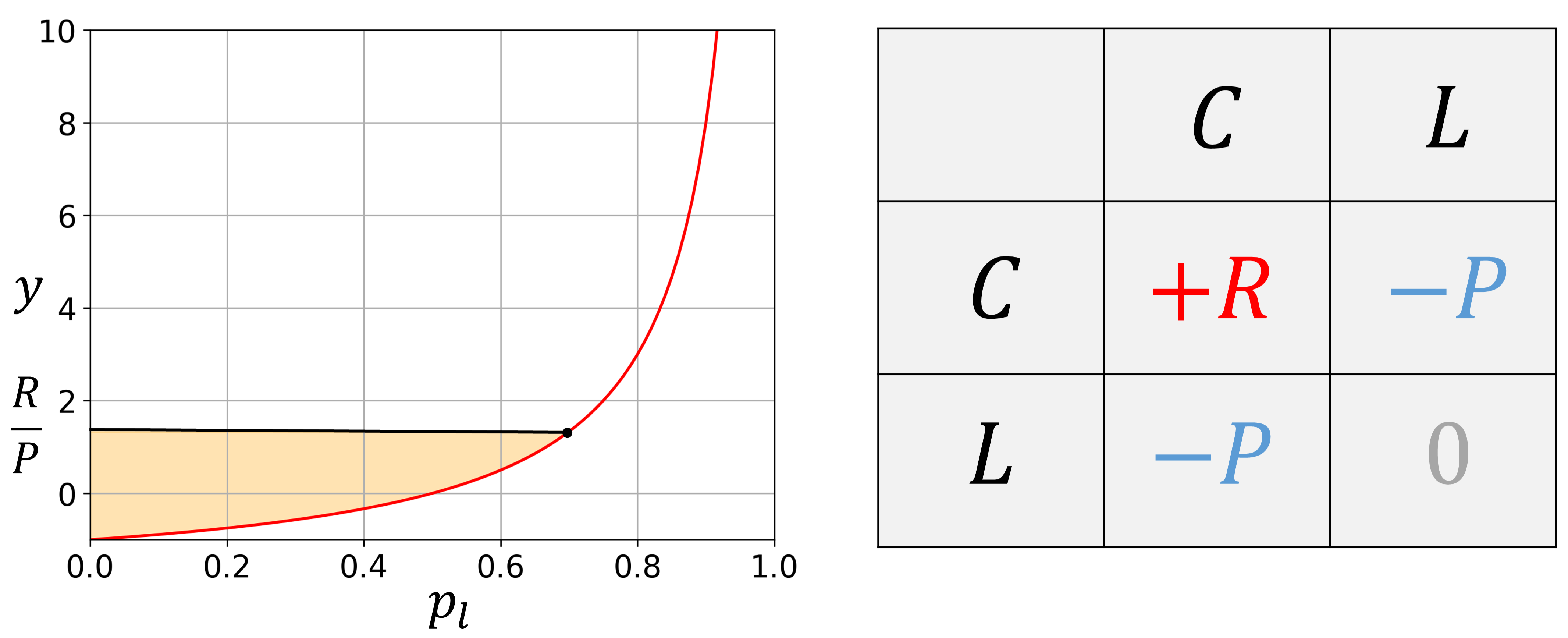}
    \caption{Left: The illustration of Eq. (\ref{eq5m}). The orange area is the payoff matrixes that can be solved by methods using $Q_i(\tau_i,a_i)$ as policy. Right: Payoff matrix of the bimanual lifting example.}
    \label{curve}
\end{figure}

In this section, we use a bimanual task in robotics as a motivating example, although our method can be applied to general tasks with with non-monotonic payoffs. 
The bimanual tasks are defined as tasks requiring a dual-arm robot to manipulate. Compared to a single-arm robot, a dual-arm robot has prominent advantages in finishing complex real-world tasks, especially humanly intuitive tasks \cite{yang2016neural, rakita2019shared}. For example, simple tasks such as lifting a single heavy object or complex tasks such as outer-space assembly and repair or domestic work \cite{lee2013relative}. However, learning the bimanual tasks using MARL is challenging since the cooperative actions between multiple arms usually depend on each other's actions. Furthermore, the payoff matrixes of these tasks are non-monotonic which prevent agents from learning the optimal policy. To illustrate the reason for such a problem, we use the bimanual lifting task as an example. 
The task requires two agents to take the cooperative action $C$ simultaneously to lift the object, and the collaborative reward is $+R$. Otherwise, the agent will receive a punishment $-P$ for taking action $C$ alone since the object cannot be lifted alone and the energy is wasted. Meanwhile, action $L$ represents lazy actions (e.g., staying still) in the action spaces that cannot achieve cooperation but avoid punishment. The corresponding non-monotonic payoff matrix is in Figure \ref{curve}. 
\label{sec1}

\begin{figure*}[t]
  \centering
  \includegraphics[width=0.75\linewidth]{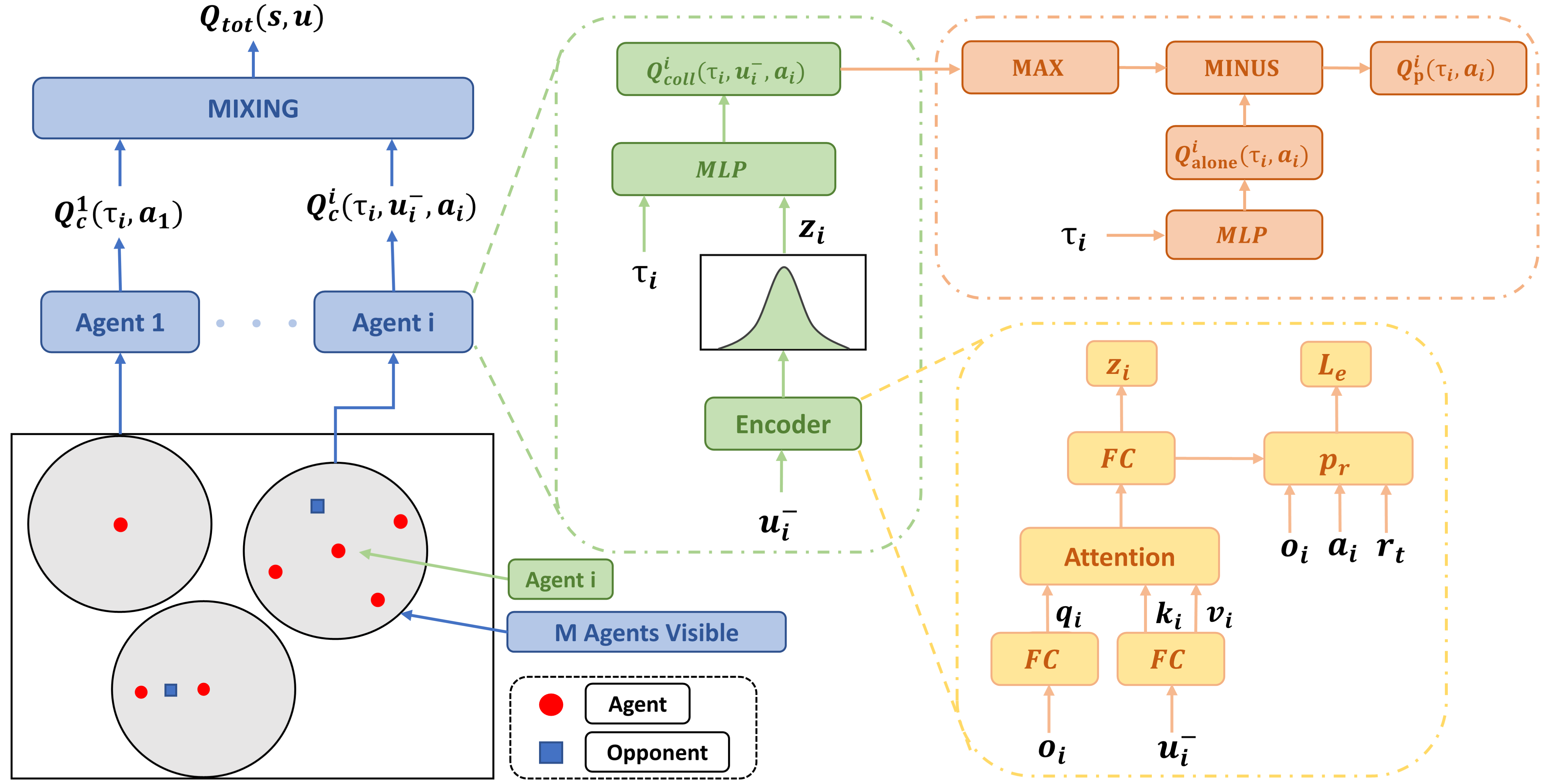}
  \caption{The architecture of our method. Left (Blue): The architecture for total Q-values. Middle (Green): The framework of the non-monotonic critic $Q_c^i(\tau_i, u_i^-, a_i)$. Upper Right (Red): The framework of greedy marginal contribution and $Q_{p}^i(\tau_i, a_i)$. Lower Right (Yellow): The framework of action encoder and predictive model.}
  \label{method}
\end{figure*}

We investigate the reason for learning difficulties in such tasks by analyzing the individual utility form. 
To learn the decentralized policy, current methods model each agent's individual utility as a certain value function $Q_i(\tau_i, a_i)$. 
However, since the returns depend on the interactions between all agents in such tasks, such a decomposition cannot consider others' actions and is insufficient to represent the optimal policy in some situations. We illustrate that a more comprehensive individual utility that can consider the interactions between agents is $Q_c^i(\tau_i, u_i^-, a_i)$ where $u_i^-$ is joint actions of potential cooperative agents (discussed in detail in Section \ref{av}). 
According to this decomposition, the individual utility $Q_i(\tau_i,a_i)$ should be viewed as a variable sampled from the distribution $Q_c^i(\tau_i, u_i^-, a_i)$ over $u_i^-$. Optimization of $Q_i(\tau_i,a_i)$ instead of $Q_c^i(\tau_i, u_i^-, a_i)$ can cause the policy converge to an average policy. We illustrate that this explains why current methods fail in environments with non-monotonic payoff matrices such as bimanual lifting. 
For instance, in the bimanual lifting task, the learned policy will not be able to represent the optimal policy when
\begin{equation}
    \begin{aligned}
        \frac{R}{P} < \frac{2p_l - 1}{1 - p_l}.
    \end{aligned}
    \label{eq5m}
\end{equation}
where $p_l$ is the probability of each policy taking action $L$.
The detailed derivation is included in Appendix 6. Figure \ref{curve} shows the result of Eq. (\ref{eq5m}). 

We find that methods directly using $Q_i(\tau_i,a_i)$ as policy such as QMIX can only solve a small part of all possible payoffs since even a small $P$ can cause the policy to fail because the weight of $P$ grows exponentially as $p_l$ grows. 
However, other methods proposed to tackle the relative overgeneralization can decrease the $p_l$ to solve more payoffs. For instance, Weighted QMIX decreases $p_l$ by placing more importance
on the better joint actions. However, since it does not model the actual interactions among agents, the importance weight is empirical. Therefore, it cannot promise the $p_l$ to decrease to zero and still fails when $P$ is large. QPLEX can reduce $p_l$ to zero by learning converged value functions with complete expressiveness capacity. However, such value functions are hard to learn as joint action spaces grow exponentially as the number of agents grows. Moreover, learning complete expressiveness requires sufficient exploration which is difficult to achieve in the example environment, since the $p_l$ is usually large during uniform exploration. This is because there are only a few cooperative actions (e.g., lift) and the lazy actions (e.g., stay and move) are major in the entire action spaces.
However, our method tackles these problems by learning a feasible value decomposition which can consider the interactions among agents and using a greedy marginal contribution to guarantee that the $p_l$ can be reduced to zero. Therefore, our method achieves both optimality and efficiency in practice. 

\section{PRELIMINARY}
\subsection{Dec-POMDP}
A fully cooperative multi-agent sequential decision-making task can be described as a decentralized partially observable Markov decision process (Dec-POMDP), which is defined by a set of possible global states $S$, actions $A_1, . . . , A_N$, and observations $\Omega_1, . . . , \Omega_N$.
At each time step, each agent $i \in \{1, ..., N\}$ chooses an action $a_i \in A_i$, and they together form a joint action $\mathbf{u} \in U$. The next state is determined by a transition function $P:S \times U \rightarrow S$. The next observation of each agent $o_i \in \Omega_i $ is updated by an observation function $O:S \rightarrow \Omega$. All agents share the same reward $r:S \times U \rightarrow \mathbf{r}$ and a joint value function $Q_{tot} = E_{s_{t+1}: \infty ,a_{t+1}: \infty } [R_t|s_t, \mathbf{u_t}]$ where $ R_t = \sum^{\infty}_{j=0} \gamma^j r_{t+j}$ is the discounted return. 
The observation of each agent can also be replaced by the history of actions and observations of each agent as a proxy to handle partial observability \cite{sunehag2017value,rashid2018qmix}. The history of actions and observations of agent $i$ can be viewed as $\tau_i$ which is $(o_i^0, a_i^0, ..., o_i^t)$. 
\subsection{Marginal Contribution}
In this work, we also introduce the marginal contribution of Shapley Value. 
The marginal contribution in Shapley Value of agent $i$ is defined as
\begin{equation}
	\begin{aligned}
        \phi_i = v(C) - v(C/i)
	\end{aligned}
    \label{eq0}
\end{equation}
where $C$ is a team consisting of agents cooperating with each other to achieve a common goal and $C/i$ represents the set with the absence of agent $i$. $v(C)$ refers to the value function for estimating the cooperation of a set of agents. Additionally, we use $N,M_i$ to refer to the total number of agents and $M_i$ is the number of agents observed by agent $i$.

\section{Method}
In this section, we propose an explicit credit assignment method, Adaptive Value decomposition with Greedy Marginal contribution (AVGM), to learn the optimal cooperative policy in tasks with non-monotonic payoffs. 
First, we propose a value decomposition which can capture the interactions between agents. Then, we propose an actor-critic structure based on the value decomposition that trains a centralized critic and calculates a greedy marginal contribution from the critic to train a decentralized execution policy.

\subsection{Learning Non-Monotonic Critic via Adaptive Value Decomposition}
\label{av}
Since current individual utility $Q_i(\tau_i, a_i)$ fails to capture the interactions between agents, we revise the agent's individual utility to a more comprehensive form. First of all, as the total rewards of environment are generated by all kinds of possible interactions between agents, we have
\begin{equation}
    \begin{aligned}
        r_{tot}(s, u) & = \sum_{i=1}^{N} \hat r_1(o_i,a_i) + \sum_{i<j=2}^{i<j=N} \hat r_{2}(o_i, o_j, a_i, a_j) \\ & + ...  + \hat r_{N}(o_1,..., o_N, a_1,..., a_N).
    \end{aligned}
    \label{eq1m}
\end{equation}
Notable, each item $\hat r_m$ in Eq. (\ref{eq1m}) exists only when there are interactions among the referred agents. Otherwise, it should be neglected as the reward is represented by other items. However, the enormous number of combinations makes it difficult to use this decomposition in practice. Fortunately, we can simplify the decomposition in multi-agent decentralized execution settings.
In decentralized execution settings, we assume the cooperative mode is that each agent should only interact with others who are observable to the agent. The reason for this is that the decentralized policy can only take action according to information within the observation. If an agent needs to cooperate with other agents that are not within its observation field, we must introduce a communication method to transfer the necessary messages.
Otherwise, the agents cannot identify which agents can potentially achieve cooperation, which would make the examples of cooperation irregular from the view of local observations and makes it impossible to learn the cooperative policy. 
Additionally, most robotics scenarios have a smaller range of interaction than the field of view in practice, which means that if an agent is within the interaction range of another agent, it must also be within its field of view.

\begin{figure*}[t]
	\centering
	\includegraphics[width=0.8\linewidth]{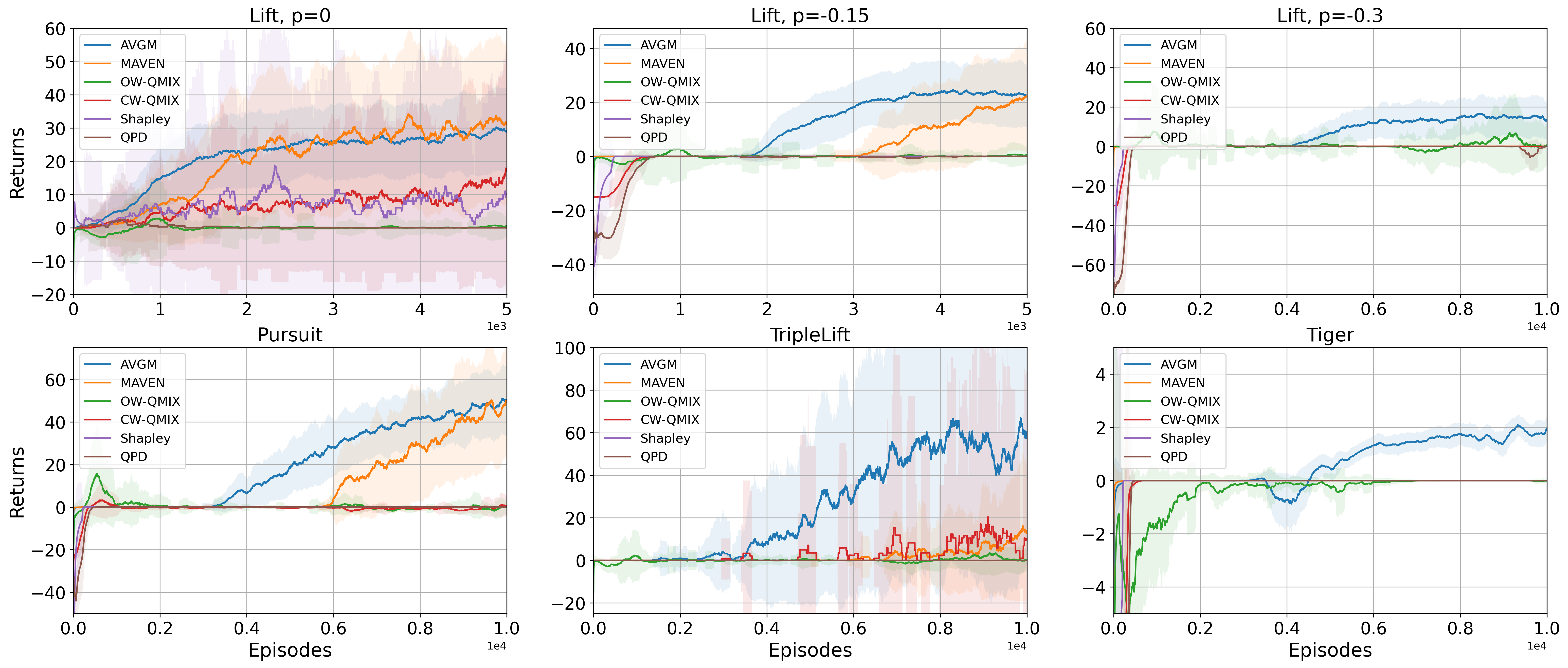}
	\caption{Top: The results in \textit{lift} with the penalty growing from 0 to -0.3. Bottom: The results in \textit{pursuit}, \textit{triplelift} and \textit{tiger}.}
	\label{rl}
\end{figure*}

Following this principle, we define team $i$ as all $M_i$ agents in the view field of agent $i$. The total reward of agent $i$ can be described as $r_{team}^i(o_i, u_i^-, a_i)$, where $u_i^-$ means the joint actions of the rest of agents in team $i$. Since observation $o_i$ can serve as the global state of the team which has included all other agents' information and agent $i$ only requires others' actions information to refer to the total reward. Furthermore, considering all kinds of possible interactions between agents, we can decompose the $r_{team}^i(o_i, u_i^-, a_i)$ as Eq. (\ref{eq1m}),
\begin{equation}
    \begin{aligned}
        r_{team}^i(o_i, u_i^-, a_i) & = \hat r_1^i(o_i, a_i) + \sum_{j=1,j \neq i}^{M_i} \hat r_2^i(o_i, a_j, a_i)
          \\ & + ... + \hat r_{M_i}^i(o_i, a_1,..., a_{M_i}, a_i).
    \end{aligned}
    \label{eq3m}
\end{equation}
where $\hat r_m^i$ exists only when there are interactions between the described agents and agent $i$, otherwise it should be equal to zero.
Combining Eq. (\ref{eq3m}) with Eq. (\ref{eq1m}), we have
\begin{equation}
    \begin{aligned}
        r_{tot}(s, u) = \sum_{i=1}^{N} r_{team}^i(o_i, u^-_i, a_i).
    \end{aligned}
    \label{eq4m}
\end{equation}
The detailed derivation can be found in Appendix 5. Based on this decomposition, we can derive the corresponding adaptive value decomposition, using $Q_c^i(\tau_i, u_i^-, a_i)$ as each agent's individual utility which learns the value of cooperating with the dynamically changing observable agents at each time step. 
Then, we have

\textbf{Theorem 1:} \textit{
    For any $r_{tot}(s, u)$, the corresponding $Q_{tot}(s,u)
            = \mathbb{E}\left[\sum_{t=0}^{\infty} \gamma^{t} r_{tot}(s, u) \mid \pi \right]$ 
    and each agent's utility $Q_c^i(\tau_i, u_i^-, a_i)$ satisfies
    \begin{equation}
        \begin{aligned}
            \mathop{\arg\max} \limits_{u}(Q_{tot}(s,u)) = & \{\mathop{\arg\max} \limits_{a_1}(Q_c^1(\tau_1, u_1^-, a_1)),..., \\ & \mathop{\arg\max} \limits_{a_N}(Q_c^N(\tau_N, u_N^-, a_N))\}.
        \end{aligned}
    \end{equation}
}
Detailed proof can be found in Appendix 5. Theorem 1 indicates that the adaptive value decomposition using utility $Q_c^i(\tau_i, u_i^-, a_i)$ can represent the unbiased value decomposition given any reward function and satisfies the IGM principle for decentralized execution. Furthermore, since the utility only involves observable agents, it does not make the learning problem harder along with the increasing of agent number within the environment.

Since we have proposed $Q_c^i(\tau_i, u_i^-, a_i; \theta)$ as an agent's utility to overcome the non-monotonic problem. 
We would like to learn a policy from it that can cooperate considering each other's actions. 
However, the policy $a_i \sim Q_c^i(\tau_i, u_i^-)$ requires $u_i^-$ to be taken but $u_i^-$ also requires $a_i$ to be produced, which is a deadlock situation. 
To address this problem, we propose an actor-critic structure that trains a centralized critic and calculates a greedy marginal contribution from the critic to train a decentralized execution policy. 

Following Theorem 1, we construct our non-monotonic centralized critic. 
Since Theorem 1 indicates that $Q_{coll}^i$ can represent the unbiased value decomposition given any reward function and satisfies the IGM principle for decentralized execution. We can use all $Q_{coll}^i$ to calculate $Q_{tot}$ through a monotonic mixing network similar to QMIX, and Theorem 1 promises the mixing value is unbiased.
Therefore, our centralized critic consists of each agent's adaptive utility $Q_c^i(\tau_i, u_i^-, a_i)$ and a mixing network to produce the global Q-value $Q_{tot}(s, u)$.
The optimization objective is mean squared error (MSE)
\begin{equation}
	\begin{aligned}
		\mathcal{L}_{TD}(\theta)=\mathbb{E}_{\pi}[Q_{tot}(s_{t}, \mathbf{u}_{t})-y_t]^2
		\\
		y_t=r_t+\gamma \max _{\mathbf{u}_{t+1}} Q_{tot}\left(s_{t+1}, \mathbf{u}_{t+1}\right).
	\end{aligned}
    \label{eq6m}
\end{equation}
However, we notice that agents are not always able to observe other agents and the value function when agents have no interactions with each other should be different. Therefore, we construct $Q_c^i(\tau_i, u_i^-, a_i)$ as two value functions,
\begin{equation}
	\begin{aligned}
        Q_c^i(\tau_i, u_i^-, a_i) = 
        \begin{cases}
            Q_{coll}^i(\tau_i, u_i^-, a_i; \theta_1) & M_i>0 \\
            Q_{alone}^i(\tau_i, a_i; \theta_2) & M_i=0
        \end{cases}
	\end{aligned}
    \label{eq7m}
\end{equation}
where $M_i$ is the number of agents observed by agent $i$. $Q_{coll}^i(\tau_i, u_i^-, a_i)$ represents the cooperative value function and is selected when any other agent can be observed. The $Q_{alone}^i(\tau_i, a_i)$ learns the selfish value function of an agent acting alone and is selected when no agents are around. During training process, we choose one of these two value functions as $Q_c^i(\tau_i, u_i^-, a_i)$ according to whether any other agent can be observed at each time step to produce $Q_{tot}(s, u)$.

\begin{figure*}[t]
	\centering
	\includegraphics[width=0.75\linewidth]{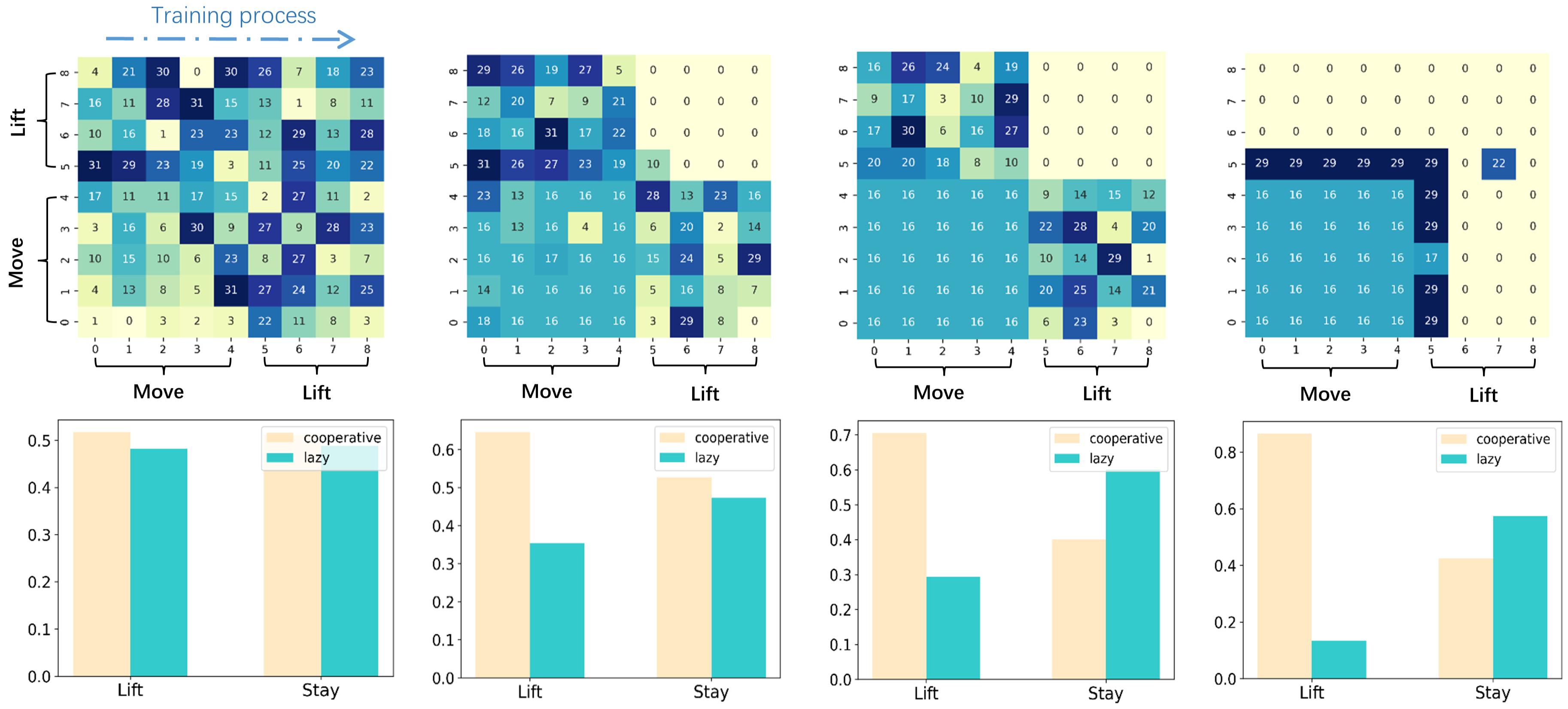}
	\caption{
        Top: The results of the action encoder outputs from initial (left) to final (right) of the training process. The x-axis and y-axis are the actions of other agents. Different colors and numbers represent different action representations. Bottom: The results of credit assignment values from initial (left) to final (right) of the training process. All results have been normalized for better representation.}
	\label{inter}
\end{figure*}

\subsection{Explicit Credit Assignment via Greedy Marginal Contribution}
Although we already have a non-monotonic critic function, we still need to learn a decentralized execution policy. 
Fortunately, we can train a decentralized policy by optimizing the marginal contribution of each agent to avoid requiring others' actions. 
The reason is that the marginal contribution can only be maximized by achieving cooperation. Therefore, if an action can maximize the marginal contribution, it is the cooperative action.
According to the definition of marginal contribution, if agent $i$ can observe other agents, we have 
\begin{equation}
    \begin{aligned}
        \phi_i(\tau_i, a_i) = 
        v(T_i) - v(T_i/i) 
        =Q_{coll}^i(\tau_i, u_i^-, a_i) - Q_{coll}^i(\tau_i, u_i^-, a_i^l)
    \end{aligned}
    \label{eq8m}
\end{equation}
where $T_i$ is the set of agents consisting of agent $i$ and all agents observed by agent $i$, $a_i^l$ is the lazy action which cannot lead to cooperation with any agent. In practice, we use 
\begin{equation}
    \begin{aligned}
        a_i^l = \mathop{\arg\max} \limits_{a_i}(Q_{alone}^i(\tau_i, a_i))
    \end{aligned}
    \label{eq9m}
\end{equation}
The rationale is that the policy represented by $Q_{alone}$ is a selfish policy only considering acting alone. In non-monotonic environments, such a policy is supposed to take lazy actions to avoid the punishment in payoff matrix. Moreover, even in monotonic environments, the cooperative actions are usually different from selfish actions, so Eq. \ref{eq9m} holds in all kinds of environments. However, there is still another problem that for any $(\tau_i, a_i)$, there are multiple $\phi_i(\tau_i, a_i)$ corresponding to different $u_i^-$. This can cause our policy to converge to an average policy like the situation we discussed in Section \ref{sec1}. To address this issue, we propose a greedy marginal contribution,
\begin{equation}
    \begin{aligned}
        \phi_i^*(\tau_i, a_i) & = Q_{coll}^i(\tau_i, u_i^{-*}, a_i) - Q_{coll}^i(\tau_i, u_i^{-*}, a_i^l) 
        \\
        (u_i^{-*}, a_i^*) & = \mathop{\arg\max} \limits_{(u_i^{-},a_i)} (Q_{coll}^i(\tau_i, u_i^-, a_i))
    \end{aligned}
    \label{eq10m}
\end{equation}
The insight is that we expect our policy to learn the potential optimal value of action $a_i$. The optimal value can be reached when all other agents take the optimal cooperative actions $u_i^{-*}$ to cooperate with agent $i$.  In other words, we encourage the agent to learn each action's value based on the optimistic belief that the possibility of other agents to take non-cooperative action is zero so that all other agents would cooperate with itself. As the main problem in the non-monotonic environment is that agents tend to take lazy actions fearing other agents not cooperating, this optimistic belief can facilitate exploration by increasing the probability of sampling cooperative actions and promises that agents will jump off the sub-optimal policy to converge to the optimal cooperative policy. 
In this way, we have our decentralized policy's value function $Q_{p}^i(\tau_i, a_i;\mu)$ and decentralized policy as $\pi(a_i|\tau_i) = \arg\max_{a_i}(Q_{p}^i(\tau_i, a_i))$. 
We use $\phi_i^*(\tau_i, a_i)$ as target when an agent can observe other agents and use $Q_{alone}$ when acting alone. The overall loss objective is 
\begin{equation}
    \begin{aligned}
        \mathcal{L}_{p}(\mu)=
        \begin{cases}
            \mathbb{E}_{\pi}[Q_{p}^i(\tau_i, a_i)-\phi_i^*(\tau_i, a_i)]^2 & M_i>0 \\
            \mathbb{E}_{\pi}[Q_{p}^i(\tau_i, a_i)-Q_{alone}(\tau_i, a_i)]^2 & M_i=0
        \end{cases}
    \end{aligned}
    \label{eq11}
\end{equation}

\subsection{Learning Actions Category Representation}
However, the structure of $Q_c^i(\tau_i, u_i^-, a_i)$ is infeasible to use in practice because $u_i^-$'s dimension is uncertain that we cannot construct a neural network for it. More importantly, we need to search for the maximization of $Q_c^i(\tau_i, u_i^-, a_i)$ when calculating our greedy marginal contribution. The time complexity will be exponential if we search all possible $u_i^-$. To address these problems, we propose an action encoder $z_i \sim f(u_i^-;\phi)$ to map the actions' information into different categories in latent space to reduce the dimension. The insight behind this is that we find many of the combinations in $u_i^-$ actually have the same effect on the environment and agent $i$, so they should have the same category in the latent space. For example, in a scenario requiring four agents to take cooperative actions, agent $i$ would face the same situation regardless of which of the other three agents takes the lazy action.

To handle the uncertain dimension of $u_i^-$, we use an attention structure \cite{graves2014neural,oh2016control} in the encoder. Moreover, the attention structure can help agents pay more attention to agents that are more likely to cooperate. We calculate the weight of each $a_j$ in $u_i^-$ according to the information of $o_i$. The overall structure of attention is shown in Figure \ref{method}. 
However, since we lack labels for the categories of the actions, the encoder has to learn in an unsupervised manner.
Therefore, in practice we use a predictive model $p_r(o_i, f(u_i^-), a_i;\eta)$ to learn the encoder, as the learned action categories should contain enough information such that the next reward of agent $i$ can be predicted when given the observations and actions of agent $i$. The overall model is trained by minimizing the following loss function which is derived from Eq. (\ref{eq4m}),
\begin{equation}
    \begin{aligned}
        \mathcal{L}_{e}(\phi, \eta)= 
        (\sum_{i=0}^{N} p_r(o_i, f(u_i^-), a_i) - r_t)^2
    \end{aligned}
    \label{eq12}
\end{equation}
where $r_t$ is the global reward. At last, we modify $Q_c^i(\tau_i, u_i^-, a_i)$ by replacing $u_i^-$ with $z_i \sim f(u_i^-)$ to $Q_c^i(\tau_i, z_i, a_i)$. This formula promises a linear search time $O(Z)$ where $Z$ is the pre-defined number of categories.

\begin{figure}[t]
	\centering
	\includegraphics[width=0.99\linewidth]{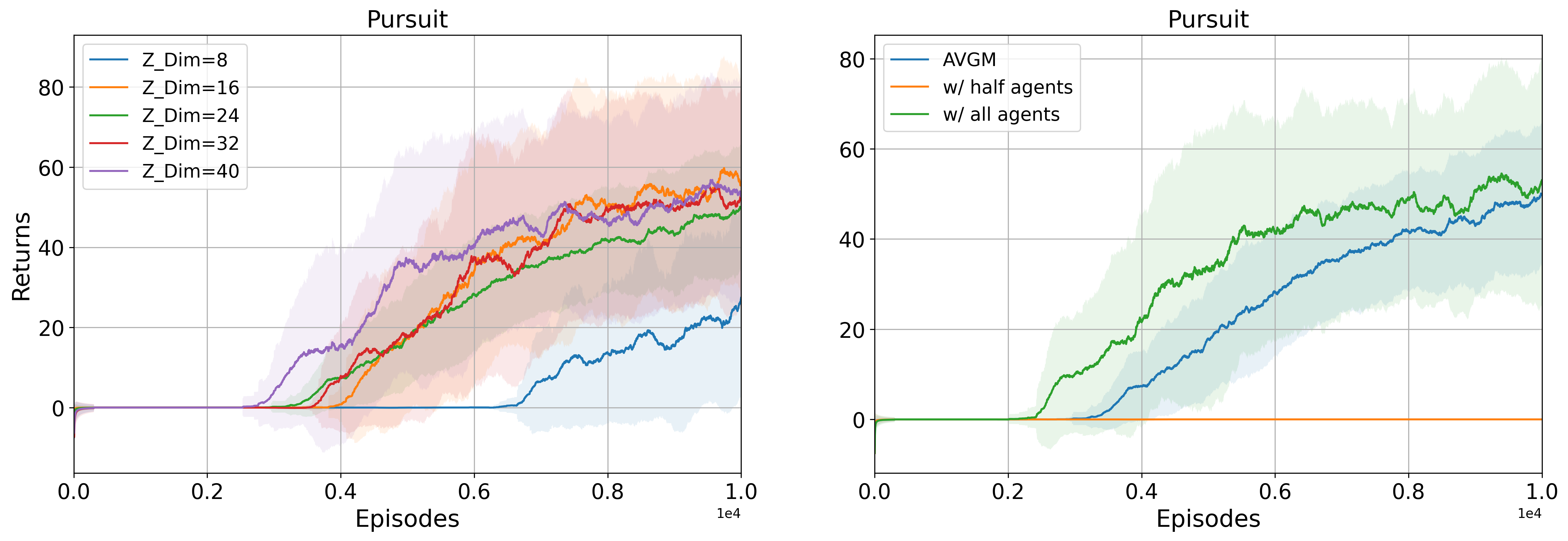}
	\caption{Left: Comparison between different $z$ dimensions. Right: Comparison between different agents' choosing methods.}
	\label{sup}
\end{figure}

\section{Experiment}
\subsection{Environments and Experimental Settings}
We design multiple challenging cooperative robotics tasks with non-monotonic payoffs to evaluate each method. The experiments are conducted based on MAgnet \cite{zheng2018magent}, where we implement four tasks with different focuses: \textit{lift}, \textit{triplelift}, \textit{pursuit} and \textit{tiger}. 
All tasks are non-monotonic environments which the tasks require all agents to take optimal cooperative actions based on others' actions. Otherwise, agents will receive penalty. \textit{lift} and \textit{triplelift} require two or more and three or more agents to lift a cargo together which are similar to the bimanual lifting task but more tricky because agents need to move to the cargo first. \textit{pursuit} is a predator-prey task that needs two or more agents to attack a prey together to capture it. \textit{tiger} is more difficult than pursuit as its prey will be killed by agent's attack actions and the prey in \textit{tiger} can recover health at each time step, which encourages agents to learn to attack prey with higher health and refrain from killing prey to get more rewards.
We use the performance of evaluation episodes with greedy action selections as the final performance. The performance is evaluated by the average return on each agent. All experiments are carried out with five random seeds. More details about the implementations of all scenarios are included in Appendix 1.

In the experiments, we compare our method AVGM with QPD, Shapley, MAVEN and WQMIX. All methods use the same basic hyperparameters and network structures with similar parameters to ensure the comparison is fair. Please refer to Appendix 2 for more experimental settings. \footnote[1]{https://github.com/Locke637/AVGM}

\begin{figure}[t]
	\centering
	\includegraphics[width=0.99\linewidth]{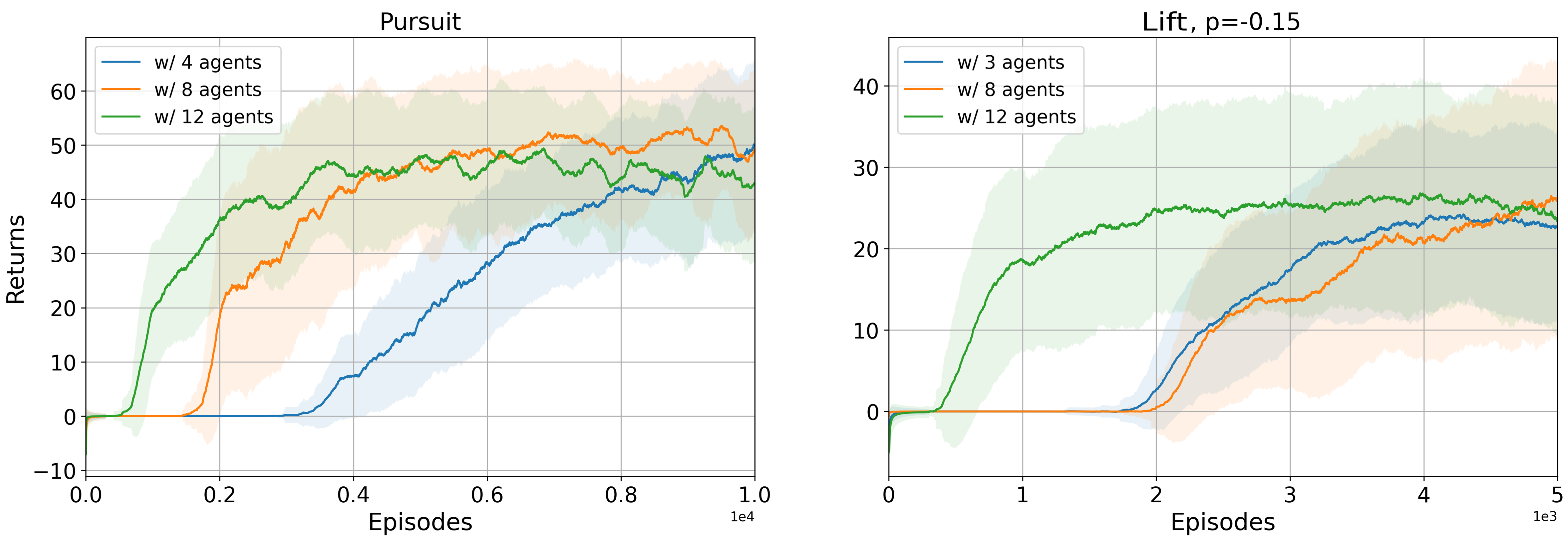}
	\caption{The results in large-scale scenarios. Left: Results in \textit{pursuit}. Right: Results in \textit{lift}.}
	\label{scale}
\end{figure}

\subsection{Performance}
We show the results of six different robotics scenarios in Figure \ref{rl}. In \textit{lift}, we demonstrate how penalties affect the learning process of all methods. We evaluate all methods in \textit{lift} with the penalty growing from 0 to -0.3. The results indicate that only AVGM can solve all scenarios while other methods' performance decreases as the penalty grows. Especially, the results show that other methods tend to learn a lazy policy that never takes cooperative actions to avoid the penalty when penalty grows larger. This phenomenon is consistent with our previous analysis in Section \ref{sec1}. 
We also notice that MAVEN and WQMIX have better performance in scenarios with larger penalties as they are designed to alleviate non-monotonic. However, they cannot solve the non-monotonic problem completely as they fail in tasks with the most significant penalty. Furthermore, the poor performance of QPD is due to the lack of considering interactions between agents, while Shapley fails in harder scenarios because the centralized critic is not sufficient to represent the true marginal contribution of optimal action, thus leading to a sub-optimal policy. In \textit{pursuit}, the result is similar as in \textit{lift} that only AVGM learns an optimal policy. In \textit{triplelift}, we find that the performance of other methods diminishes as cooperation becomes harder to achieve. However, AVGM still solves the task by considering others' actions and searching for the optimal marginal contribution. Finally, all other methods fail in \textit{tiger} as inefficient exploration would kill all prey in this task, making it difficult to find a cooperative mode. The result indicates that AVGM is highly efficient in learning in non-monotonic environments.

\subsection{Analysis of Interpretability}
In this section, we analyze several properties of our method to show the interpretability of AVGM. To better demonstrate, we choose a typical situation in \textit{lift} with penalty -0.3, where all agents are in the right position and ready to lift the cargo. The schematic diagram is included in Appendix 3. This situation is suitable for demonstrating interpretability as agents must consider each other's actions to decide whether to take cooperative or lazy actions.


In the top of Figure \ref{inter}, we demonstrate the output changes of the action encoder during training. We illustrate that the action encoder learns a meaningful category representation of others' actions. The representation classifies others' actions according to whether they take lift actions or movement actions. We notice that the encoder first learns the easy situations that both agents are lifting or moving. Through further training, the encoder learns to classify the rest combinations of actions.

Furthermore, we show the interpretability of our marginal contribution at the bottom. We compare the marginal contribution of taking optimal lift action and lazy stay action under the condition that others are taking cooperative actions or lazy actions. The result shows that the marginal contribution of taking lift action has an increasingly large advantage given other's lift actions over given other's lazy actions as the training progresses.
Meanwhile, the marginal contribution of taking stay action when others are taking lazy actions is larger than that of others taking cooperative actions. The results show that the marginal contribution is reasonable as it is consistent with the analysis using prior knowledge. Therefore, we showcase the interpretability of our method which guarantees further potential applicants in the real-world scenarios since we can explain the behavior of each agent and reduce the unpredictable risks. 
Finally, we also demonstrate the variance of all agents' $Q_c^i(\tau_i, u_i^{-*}, a_i^*)$ during training, which shows that the agents would learn a consistent perception of the environment when considering others' actions in Appendix 4.
\begin{figure}[t]
	\centering
	\includegraphics[width=0.95\linewidth]{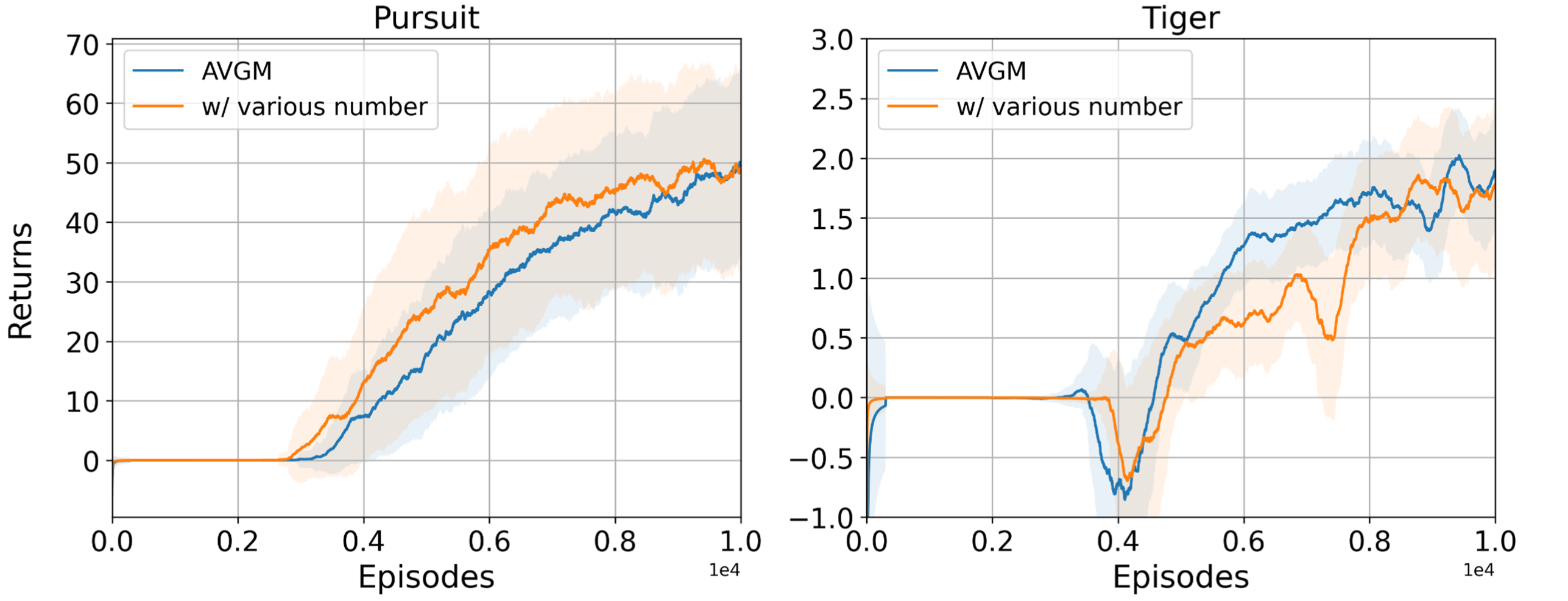}
	\caption{Results of scalability. Left: Results in \textit{pursuit}. Right: Results in \textit{tiger}.}
	\label{abl2}
\end{figure}
\subsection{Ablations}
\subsubsection{Comparison among Using Different $z$ Dimensions}
We compare the performance of AVGM using different $z$ dimensions (8, 16, 24, 32, 40) in \textit{pursuit}. The results are shown in Figure \ref{sup}. The results indicate that using different $z$ dimensions has a limited impact on the final performance when the number of dimensions is over 16. However, using 8-dimension $z$ can decrease the performance significantly as such low-dimensional latent action embedding is insufficient to represent all the necessary behavior modes. This showcases the robustness of AVGM over the hyperparameter of $z$ dimensions within a reasonable range.

\subsubsection{Comparison among Using Different Agents' Choosing Methods}
We compare the performance of using all agents, using only visible agents and using part of visible agents in \textit{pursuit}. The results are shown in Figure \ref{sup}. Specifically, using part of visible agents means that we randomly abandon half of the visible agents. The results show that using all agents and only visible agents has similar performance. This indicates that the information among visible agents is enough to represent the optimal policy, which is in line with our analysis. Meanwhile, the performance of using part of visible agents decreases significantly. 
This indicates that the absence of the joint actions of visible agents $u_i^-$ leads to a suboptimal policy.

\subsubsection{Scalability}
First, we compare the performance in large-scale scenarios with more agents (4, 8, 12) in \textit{pursuit} and (3, 8, 12) in \textit{lift}. The results are shown in Figure \ref{scale}. The results show that the final performance is similar in all settings. This indicates that adding agents has limited influence on the performance of AVGM, as the number of agents in the view of each agent is much smaller and limited compared with the total number of all agents in the environment. This demonstrates that AVGM can learn in large-scale scenarios that are intractable for methods that learn value functions with complete expressiveness. 
Moreover, training with more agents can learn cooperation at the early stages since the policy is trained with more experience and the odds of discovering the cooperative mode become higher. 
Additionally, since the structure of AVGM can handle variable input dimensions and the policy can learn to cooperate with scalable agents, we investigate whether the learned policy can be zero-shot transferred across scenarios with different numbers of agents.
We test the scalability of our method in \textit{pursuit} and \textit{tiger} with a varying agents' number of 3-5. The results in Figure \ref{abl2} indicate that our method can generalize to scenarios with different numbers of agents because there is no significant performance degradation in the test environments. 

\begin{figure}[t]
	\centering
	\includegraphics[width=0.9\linewidth]{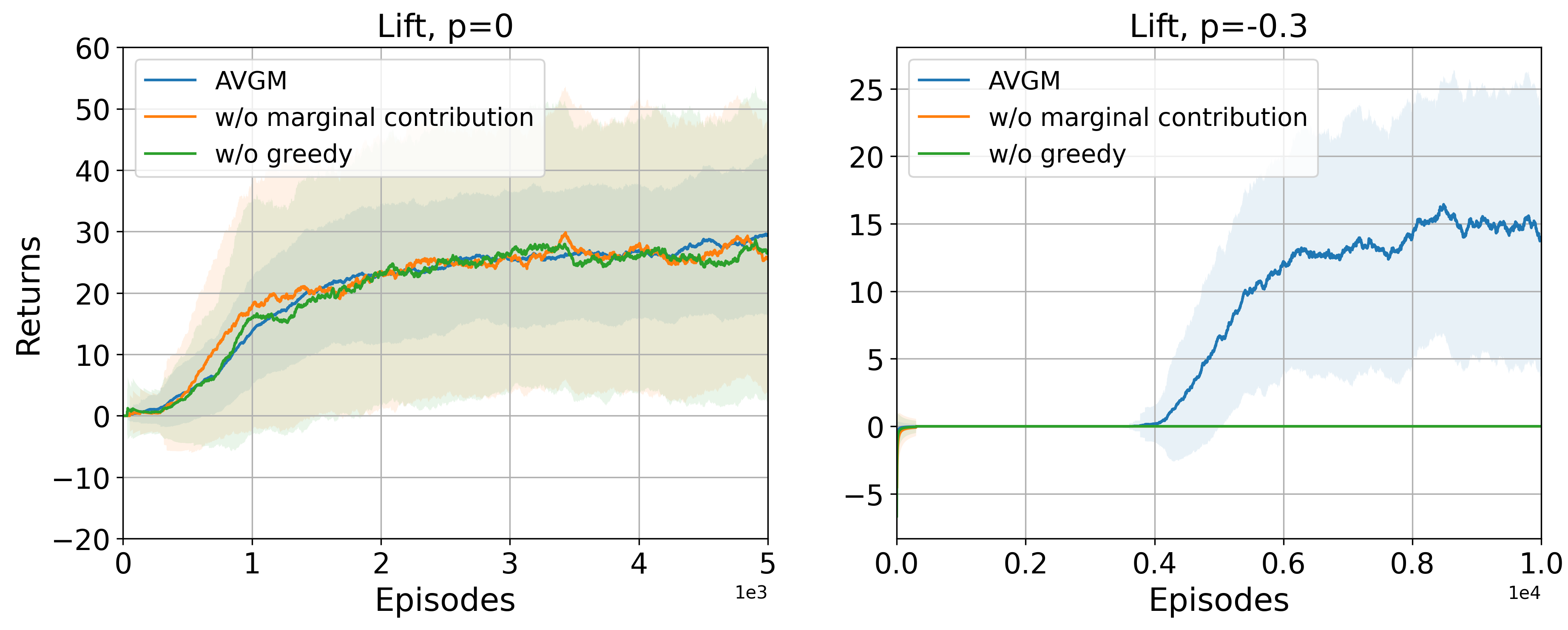}
	\caption{Results of ablation studies for marginal contribution. Left: Results in \textit{lift} with penalty of 0. Right: Results in \textit{lift} with penalty of -0.3.}
	\label{abl1}
\end{figure}

\subsubsection{Ablations for Marginal Contribution} We also conduct ablation studies on the marginal contribution, including using the actual marginal contribution instead of the greedy one and not using marginal contribution by optimizing policy directly through using $Q_c^i(\tau_i, u_i^{-*}, a_i)$. The ablation studies are conducted in \textit{lift} with penalties of 0 and -0.3. 
The results are shown in Figure \ref{abl1}. 
We find that lacking these parts of our method leads to significant reductions in non-monotonic scenarios, while the influence is limited in monotonic ones.
\section{Conclusion}
In this work, we propose a novel explicit credit assignment method to solve the robotics tasks with non-monotonic returns. 
Our method, AVGM, is based on an adaptive value decomposition considering other observable agents' actions and uses greedy marginal contribution to encourage agents to learn the optimal cooperative policy. 
This enables our method to learn cooperation in tasks with non-monotonic returns, which is a challenge for existing value-based CTDE methods.
Moreover, we propose an action encoder to guarantee that the time complexity of AVGM is linear. 
The experiments show significant performance improvement and the interpretability of AVGM.

\balance

\section{Acknowledgments}
The publication was supported by NSFC 62088101 Autonomous Intelligent Unmanned Systems and by a Grant from The National Natural Science Foundation of China (No. U21A20484).


\bibliographystyle{ACM-Reference-Format} 
\bibliography{reference}

\appendix
\onecolumn

\section{Scenarios Settings and Training Details}
\label{scen}
In MAgent, each agent corresponds to one grid and has a local observation that contains a square view centered at the agent and a feature vector including coordinates, health point (HP) and ID of agents nearby, and the agent's last action. The discrete actions are moving, staying, attacking. 
We choose four different scenarios \textit{lift}, \textit{triplelift}, \textit{pursuit} and \textit{tiger}. 
\textit{lift} and \textit{triplelift} require two or more and three or more agents to lift a cargo together which are similar to the bimanual lifting task but more difficult because agents need to move to the cargo first. \textit{pursuit} is a predator-prey task that needs two or more agents to attack a prey together to capture it. \textit{tiger} is more difficult that prey will be killed by agent's attack actions and the prey in \textit{tiger} can recover health at each time step, which encourages agents to learn to attack prey with higher health and refrain from killing prey to get more rewards.
There are the detailed settings of these scenarios, as shown in Table \ref{table2}. We demonstrate the payoff matrix by showing the $R$ as reward returned when cooperation achieved and $P$ are penalty when taking cooperative action but fail to achieve cooperation. The global state of MAgent is a mini-map ($6 \times 6$) of the global information. The opponent's policies used in experiments are randomly escaping policy in \textit{pursuit} and \textit{tiger}. 

\begin{table*}[h]
	\centering
	\scalebox{0.95}{
	\begin{tabular}{|c|c|c|c|c|c|c|}
	\hline & Lift & Pursuit & TripleLift & Tiger \\
	\hline Agent number & 3 & 4 & 3 & 4  \\
	\hline Enemy number & 3 & 4 & 3 & 4  \\
	\hline Map size & 6 $\times$ 6 & 7 $\times$ 7 & 6 $\times$ 6 & 7 $\times$ 7 \\
	\hline Patoff & R=1,P=-0.3 & R=1.5,P=-0.3 & R=2,P=-0.005 &R=1.5,P=-0.3  \\
	\hline
	\end{tabular}
	}
	\caption{Settings of MAgent Scenarios.}
	\label{table2}
\end{table*}

\begin{table}[h]
	\centering
	\begin{tabular}{|c|c|c|c|c|}
	\hline 0 & 0 & 0 & 0& 0\\
	\hline 0 & 0 & A & 0& 0 \\
        \hline 0 & A & C & A& 0 \\
        \hline 0 & 0 & 0 & 0& 0 \\
	\hline 0 & 0 & 0 & 0& 0 \\
	\hline
	\end{tabular}
    \caption{Schematic Diagram}
    \label{t3}
\end{table}

We set the discount factor as 0.99 and use the RMSprop optimizer with a learning rate of 5e-4. The $\epsilon$-greedy is used for exploration with $\epsilon$ annealed linearly from 1.0 to 0.05 in 700k steps.
The batch size is 100 and updating the target every 200 episodes. The length of each episode in MAgent is limited to 100 step. We run all the experiments five times with different random seeds and plot the mean/std in all the figures. All experiments are carried out on the same computer, equipped with an Intel i7-7700K, 64GB RAM and an NVIDIA GTX3090. The system is Ubuntu 18.04 and the framework is PyTorch. 

\section{Details of Model Implementation and Hyperparameters}
\label{model}
The network of all compared methods uses the same LSTM network, consisting of a recurrent layer comprised of a GRU with a 64-dimensional hidden state, with one fully-connected layer before and two after. All mixing networks use a fully-connected layer with 32-dimensional hidden state. 
The network of our critic and policy uses two fully-connected layers with 64-dimensional hidden state and one fully-connected layers with 32-dimensional hidden state after. The action encoder network uses one fully-connected layer with 64-dimensional hidden state and one fully-connected layer with 64-dimensional hidden state to output the category result. The category latent space's size is 32 in all experiments. The predictive model uses two fully-connected layers with 64-dimensional hidden state and one fully-connected layers with 32-dimensional hidden state after.
\section{Schematic Diagram of Section Analysis of Interpretability}
The example we used in experiments is shown in Table \ref{t3},
where $A$ in the tables means the agents, 0 means empty space and $C$ means cargo. This situation is when all agents are in the right position and ready to lift the cargo. However, agents must consider each others' actions to decide whether to take cooperative actions to finish the cooperation task or lazy actions to avoid punishment.

\section{Analysis of Variance of All Agents' Individual Utilities}
\begin{figure}[t]
	\centering
	\includegraphics[width=0.6\linewidth]{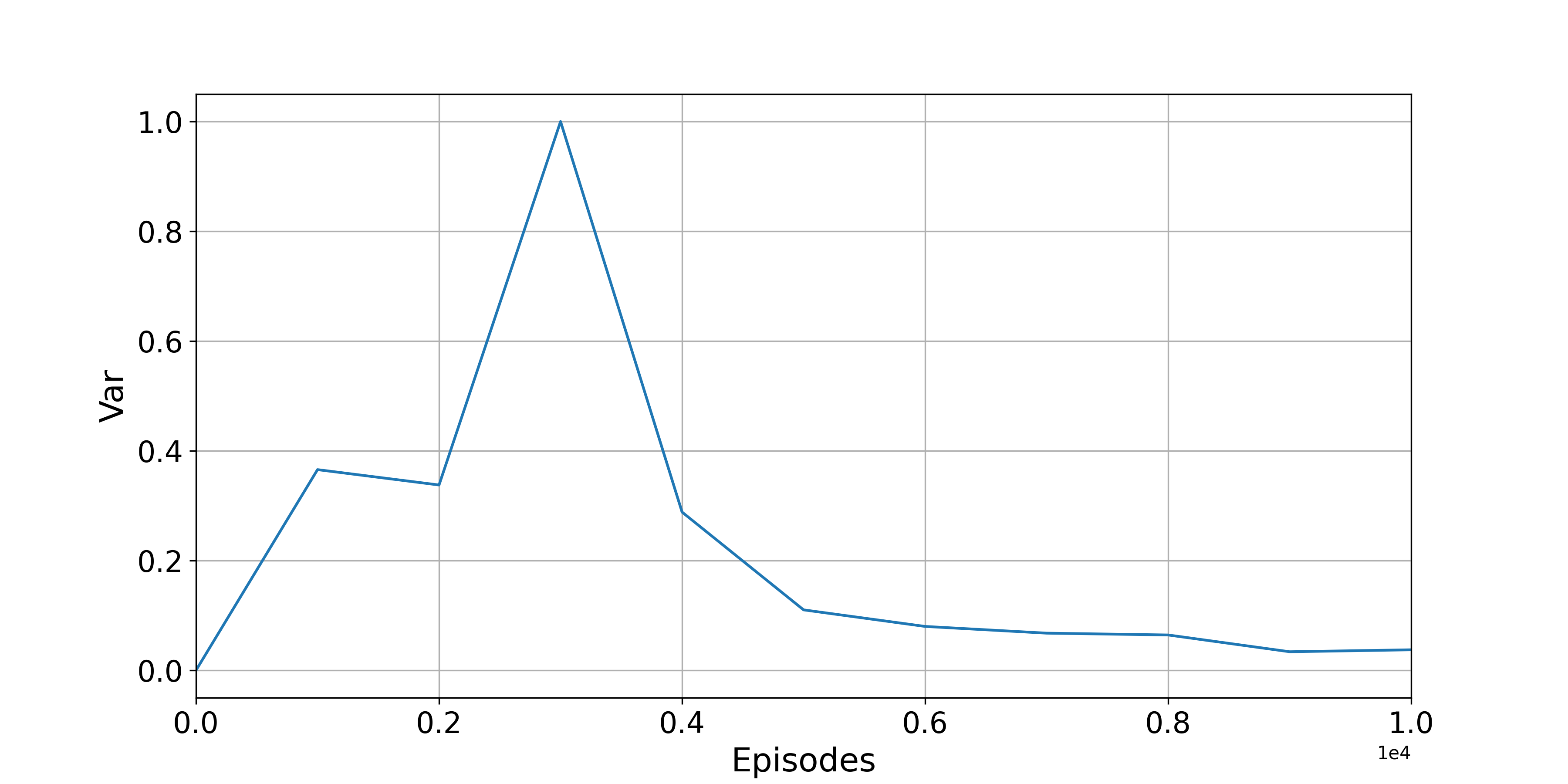}
	\caption{The result of variance of all agents' individual utilities during the training process.}
	\label{var}
\end{figure}
We demonstrate the variance of all agents' $Q_c^i(\tau_i, u_i^{-*}, a_i^*)$ during training in Figure \ref{var}. This value means the Q-value of the optimal cooperative action of each agent when given the belief that all others will take the cooperative actions. According to our method, the variance should become smaller as training processes. The reason is that agents would learn a consistent perception of the environment when considering others' actions. The result is in line with our conclusion. Notably, the variance rises at the beginning of training is because learning of value function can increase the variance compared to the initial randomly generated value. This result proves that the adaptive individual utility can learn the correct value of cooperation which is the fundamental of explicit credit assignment. 

\section{Proof of The Value Decomposition}
First of all, as the total rewards of environment are generated by all kinds of possible interactions between agents, we have
\begin{equation}
    \begin{array}{c}
        r_{tot}(s, u) = \sum_{i=1}^{N} \hat r_1(o_i,a_i) + \sum_{i<j=2}^{i<j=N} \hat r_{2}(o_i, o_j, a_i, a_j) + ... + \hat r_{N}(o_1,..., o_N, a_1,..., a_N).
    \end{array}
    \label{eq1}
\end{equation}



And we assume the cooperative mode is that every agent should only have interactions with the agents who can be observed by themselves in decentralized execution settings.

Following this principle, we define $r_{team}^i(o_i, u_i^-, a_i)$ and decompose it as,
\begin{equation}
    \begin{array}{c}
        r_{team}^i(o_i, u_i^-, a_i) = \hat r_1^i(o_i, a_i) + \sum_{j=1,j \neq i}^{M_i} \hat r_2^i(o_i, a_j, a_i)
          + ... + \hat r_{M_i}^i(o_i, a_1,..., a_{M_i}, a_i).
    \end{array}
    \label{eq3}
\end{equation}
We notice that for $\hat r_m^i = \hat r_m^i(o_i, a_j,...a_m, a_i)$ if $\hat r_m^i$ is not zero, we have 
\begin{equation}
    \begin{array}{c}
        \hat r_m^i(o_i, a_j,...a_m, a_i) = \frac{1}{m} \hat r_m(o_i, o_j,...,o_m, a_j,...a_m, a_i).
    \end{array}
    \label{eq4}
\end{equation}
The reason is that if $\hat r_m$ exists, it represents the cooperation reached by agents set $(i,j, .., m; j<i<m)$. And as the specific joint action is taken by all agents in the set, each agent should equally share the cooperative reward $\hat r_m$. Then, we can modify Eq. (\ref{eq3}) as
\begin{equation}
    \begin{array}{c}
        r_{team}^i(o_i, u_i^-, a_i) = \hat r_1(o_i, a_i) +  \frac{1}{2} \sum_{j=1,j  \neq i}^{M_i} \hat r_2(o_i,o_j, a_j, a_i)
          + ... \\ + \frac{1}{M_i} \hat r_{M_i}(o_i,o_1,...,o_{M_i}, a_1,..., a_{M_i}, a_i).
    \end{array}
    \label{eq5}
\end{equation}
We sum all $r_{team}^i(o_i, u_i^-, a_i)$
\begin{equation}
    \begin{array}{c}
        \sum_{i=1}^{N} r_{team}^i(o_i, u^-_i, a_i) = \sum_{i=1}^{N} \hat r_1(o_i, a_i) +  \sum_{i=1}^{N} \frac{1}{2} \sum_{j=1,j  \neq i}^{M_i} \hat r_2(o_i,o_j, a_j, a_i)
          + ... 
		  \\=\sum_{i=1}^{N} \hat r_1(o_i, a_i)+ \frac{1}{2} \sum_{i=1}^{N} \sum_{j=1,j  \neq i}^{M_i} \hat r_2(o_i,o_j, a_j, a_i) + \sum_{i=1}^{N} \sum_{j=1,j  \neq i}^{N-M_i} 0
          + ... 
    \end{array}
    \label{eqq5}
\end{equation}
where 0 stands for the agents out of view of agent $i$ and the corresponding reward is zero. Then, we have 
\begin{equation}
    \begin{array}{c}
		\\\sum_{i=1}^{N} \hat r_1(o_i, a_i)+ \frac{1}{2} \sum_{i=1}^{N} \sum_{j=1,j  \neq i}^{M_i} \hat r_2(o_i,o_j, a_j, a_i) + \sum_{i=1}^{N} \sum_{j=1,j  \neq i}^{N-M_i} 0
		+ ... 
		  \\ = \sum_{i=1}^{N} \hat r_1(o_i, a_i)+ \frac{1}{2} \sum_{i=1}^{N} \sum_{j=1,j  \neq i}^{N} \hat r_2(o_i,o_j, a_j, a_i) + ... 
		\\ = \sum_{i=1}^{N} \hat r_1(o_i, a_i)+ \frac{1}{2} \sum_{i<j=2}^{i<j=N} 2 \times \hat r_2(o_i,o_j, a_j, a_i) + ... 
		\\ = \sum_{i=1}^{N} \hat r_1(o_i, a_i)+ \sum_{i<j=2}^{i<j=N} \hat r_2(o_i,o_j, a_j, a_i) + ... \\ = r_{tot}(s, u).
    \end{array}
\end{equation}
So, we have 
\begin{equation}
    \begin{array}{c}
        r_{tot}(s, u) = \sum_{i=1}^{N} r_{team}^i(o_i, u^-_i, a_i).
    \end{array}
    \label{eq6}
\end{equation}
Then, we define the value decomposition $Q_c^i$ which models each agent's individual utility. From Eq. (\ref{eq6}), we have
\begin{equation}
	\begin{array}{c}
		Q_{tot}(s,u) =\mathbb{E}\left[\sum_{t=0}^{\infty} \gamma^{t} r_{tot}\left({s}, u\right) \mid \pi \right] 
		= \mathbb{E}\left[\sum_{t=0}^{\infty} \gamma^{t} \sum_{i=1}^{N} r_{team}^i(o_i, u^-_i, a_i) \mid \pi \right]
        =\sum_{i=1}^{N} Q_c^i(s,u).
	\end{array}
    \label{eq77}
\end{equation}
In addition, we have
\begin{equation}
    \begin{array}{c}
        \mathop{\arg\max} \limits_{a_i}(Q_{tot}(s,u)) = \mathop{\arg\max} \limits_{a_i}(Q_c^i(s,u)) = \mathop{\arg\max} \limits_{a_i}(Q_c^i(\tau_i, u_i^-, a_i)).
    \end{array}
    \label{eq8}
\end{equation}
The first part is because the value of $a_i$ is represented by item $Q_c^i$ and the reason for the second part is that $Q_c^i$ is only related to agent $i$ and all agents that can be observed by agent $i$, and all their necessary information is contained in $(\tau_i, u_i^-, a_i)$, so we can get the unbiased estimated value of  $Q_c^i$ given $(\tau_i, u_i^-, a_i)$. 
Therefore, from Eq. \ref{eq77} and Eq. \ref{eq8} we have
\begin{equation}
    \begin{array}{c}
        \mathop{\arg\max} \limits_{u}(Q_{tot}(s,u)) = \{\mathop{\arg\max} \limits_{a_1}(Q_c^1(\tau_1, u_1^-, a_1)),...,\mathop{\arg\max} \limits_{a_N}(Q_c^N(\tau_N, u_N^-, a_N))\}.
    \end{array}
    \label{eq9}
\end{equation}
An intuitive understanding of Eq. \ref{eq9} is that each agent takes action based on the perception of other cooperative agents' actions, so them can take the corresponding cooperative action and the joint action is the optimal cooperative joint action.
In conclusion, Eq. \ref{eq9} indicates that the utility $Q_c^i(\tau_i, u_i^-, a_i)$ can satisfy the IGM principle given ground true $Q_{tot}(s,u)$ instead of the monotonic centralized value function learned by other methods such as QMIX. This proves that the adaptive value decomposition using utility $Q_c^i(\tau_i, u_i^-, a_i)$ can represent the unbiased value decomposition given any reward function and satisfies the IGM principle for decentralized execution. 
Therefore, we have proved our Theorem 1.


\section{Analysis of Limitation of Individual Utility}
We analyze the limitation of individual utility $Q_i(\tau_i,a_i)$ given the non-monotonic payoff matrix in Table \ref{t1}. 
\begin{table}
    \centering
    \begin{tabular}{|c|c|c|}
        \hline   & C  & L  \\
        \hline C & +R & -P \\
        \hline L & -P & 0  \\
        \hline
    \end{tabular}
    \caption{Non-monotonic payoff matrix}
    \label{t1}
\end{table}
We indicate that the individual utility $Q_i(\tau_i,a_i)$, should be viewed as a variable sampled from distribution $Q_c^i(\tau_i, u_i^-, a_i)$. Following this conclusion, we have the loss of $Q_i(\tau_i,a_i)$ should be 
\begin{equation}
    \begin{array}{c}
        \mathcal{L}_i = \sum_{k=1}^{K_i} p_k \cdot( \hat Q_c^{i}(\tau_i, u_i^{k-}, a_i)-Q_i(\tau_i,a_i))^2.
    \end{array}
\end{equation}
where $\hat Q_c^{i}$ means the ground true value function, $u_i^{k-}$ means one of the combination of $u_i^{-}$ and $p_k$ is the possibility of $u_i^{k-}$ occurred. 
Therefore, $Q_i(\tau_i,a_i)$ learns to the converged value by optimizing $L_i$, we have the converged $\hat Q_i(\tau_i,a_i)$ when $L_i$ is minimized,
\begin{equation}
    \begin{array}{c}
        \hat Q_i(\tau_i,a_i) = \sum_{k=1}^{K_i} p_k \cdot \hat Q_c^{i}(\tau_i, u_i^{k-}, a_i).
    \end{array}
\end{equation}
In non-monotonic environments, there is usually one specific joint action can lead to cooperation and anyone who takes a lazy action can cause failure of the tasks. In this way, we have the value of cooperative action $a_i^*$ as 
\begin{equation}
    \begin{array}{c}
        \hat Q_i(\tau_i,a_i^*) = \sum_{c=1}^{C_i} p_c \cdot \hat Q_c^{i}(\tau_i, u_i^{c-*}, a_i^*) + \sum_{l=1}^{L_i} p_l \cdot \hat Q_c^{i}(\tau_i, u_i^{l-}, a_i^*).
    \end{array}
\end{equation}
where $p_c$ means the possibility of other agent taking cooperative actions $u_i^{c-*}$ and $p_l$ means the possibility of other agents taking lazy actions $u_i^{l-}$. Additionally, we have 
\begin{equation}
    \begin{array}{c}
        \sum_{c=1}^{C_i} p_c + \sum_{l=1}^{L_i} p_l = 1
    \end{array}
\end{equation}
Similarly, we have the value of lazy action $a_i^-$ as
\begin{equation}
    \begin{array}{c}
        \hat Q_i(\tau_i,a_i^-) = \sum_{c=1}^{C_i} p_c \cdot \hat Q_c^{i}(\tau_i, u_i^{c-*}, a_i^-) + \sum_{l=1}^{L_i} p_l \cdot \hat Q_c^{i}(\tau_i, u_i^{l-}, a_i^-).
    \end{array}
\end{equation}
We know the policy represented by $Q_i(\tau_i,a_i)$ fails when $\hat Q_i(\tau_i,a_i^-)$ is larger than $\hat Q_i(\tau_i,a_i^*)$, which is 
\begin{equation}
    \begin{array}{c}
        \hat Q_i(\tau_i,a_i^-) - \hat Q_i(\tau_i,a_i^*) = \sum_{c=1}^{C_i} p_c \cdot (\hat Q_c^{i}(\tau_i, u_i^{c-*}, a_i^-)-Q_c^{i}(\tau_i, u_i^{c-*}, a_i^*)) \\+ \sum_{l=1}^{L_i} p_l \cdot (\hat Q_c^{i}(\tau_i, u_i^{l-}, a_i^-) - Q_c^{i}(\tau_i, u_i^{l-}, a_i^*)) > 0
    \end{array}
    \label{eq7}
\end{equation}
We take the example payoff matrix into Eq. (\ref{eq7}),
\begin{equation}
    \begin{array}{c}
        \hat Q_i(\tau_i,a_i^-) - \hat Q_i(\tau_i,a_i^*) = p_c \cdot (-P-R) + p_l \cdot (0 - (-P)) 
        = (p_l - 1) \cdot (P+R) + p_l \cdot P
        > 0
    \end{array}
\end{equation}
This means the policy represented by $Q_i(\tau_i,a_i)$ will fail when 
\begin{equation}
    \begin{array}{c}
        R \cdot (1 - p_l) < (2p_l - 1) \cdot P.
    \end{array}
\end{equation}
which equals to
\begin{equation}
    \begin{array}{c}
        \frac{R}{P} < \frac{2p_l - 1}{1 - p_l}.
    \end{array}
\end{equation}

\end{document}